\documentclass[letterpaper, 10 pt, conference]{ieeeconf}  % Comment this line out if you need a4paper

\IEEEoverridecommandlockouts                              % This command is only needed if 
\usepackage{graphics} % for pdf, bitmapped graphics files
\usepackage{epsfig} % for postscript graphics files
\usepackage{mathptmx} % assumes new font selection scheme installed
\usepackage{times} % assumes new font selection scheme installed
\usepackage{amsmath} % assumes amsmath package installed
\usepackage{amssymb}  % assumes amsmath package installed
\usepackage{hyperref}

\usepackage[backend=biber,style=ieee,natbib=true]{biblatex} 
\usepackage{float}
\usepackage{etoolbox}
\usepackage[dvipsnames]{xcolor}
\usepackage[margin=0.751in]{geometry}

\makeatletter
\patchcmd{\@makecaption}
  {\scshape}
  {}
  {}
  {}
\makeatletter
\patchcmd{\@makecaption}
  {\\}
  {.\ }
  {}
  {}
\makeatother

\usepackage{array}
\usepackage{colortbl}

\title{\LARGE \bf Sim-to-Real Learning of All Common Bipedal Gaits \\ via Periodic Reward Composition}
\author{\authorblockN{Jonah Siekmann$^*$\thanks{* denotes equal contribution, order determined by coin toss}, Yesh Godse$^*$, Alan Fern, Jonathan Hurst}
\authorblockA{Collaborative Robotics and Intelligent Systems Institute\\
    Oregon State University\\
    \emph{\{siekmanj, godsey, afern, jhurst\}@oregonstate.edu}}
}
\begin{document}

\maketitle

\thispagestyle{empty}
\pagestyle{empty}

\begin{abstract}
We study the problem of realizing the full spectrum of bipedal locomotion on a real robot with sim-to-real reinforcement learning (RL).
A key challenge of learning legged locomotion is describing different gaits, via reward functions, in a way that is intuitive for the designer and specific enough to reliably learn the gait across different initial random seeds or hyperparameters.
A common approach is to use reference motions (e.g. trajectories of joint positions) to guide learning.
However, finding high-quality reference motions can be difficult and the trajectories themselves narrowly constrain the space of learned motion.
At the other extreme, reference-free reward functions are often underspecified (e.g. move forward) leading to massive variance in policy behavior, or are the product of significant reward-shaping via trial-and-error, making them exclusive to specific gaits.
In this work, we propose a reward-specification framework based on composing simple probabilistic periodic costs on basic forces and velocities.
We instantiate this framework to define a parametric reward function with intuitive settings for all common bipedal gaits - standing, walking, hopping, running, and skipping.
Using this function we demonstrate successful sim-to-real transfer of the learned gaits to the bipedal robot Cassie, as well as a generic policy that can transition between all of the two-beat gaits.
\end{abstract}

\section{Introduction}

% Training a neural network to perform legged locomotion in a way which is physically feasible and robust to disturbances is an unsolved problem.
% Oftentimes, motion capture (sometimes referred to as a reference trajectory in reinforcement learning) is used to guide learning \cite{Zhang2018, Starke2020localmotion, xie2019iterative, Siekmann2020, Peng2018deepmimic} along a path which is already known to be feasible.
% This can be limiting, as often it would be preferred to learn a continuum of behaviors rather than to extrapolate from a single reference or attempt to interpolate between several reference motions.
% In addition, adherence to the reference trajectory becomes an additional constraint at the cost of preventing the policy from learning behaviors which could be more robust or efficient.
% Training a policy to perform any sort of dynamical task involves learning to react appropriately and efficiently to a massive space of possible interactions with the world; reference trajectories capture only a tiny subsection of this space, resulting in brittle behaviors.

Using reinforcement learning (RL) to learn all of the common bipedal gaits found in nature for a real robot is an unsolved problem. A key challenge of learning a specific locomotion gait via RL is to communicate the gait behavior through the reward function. In general, a specific gait can be viewed as a dynamic process that has a characteristic periodic structure, but is also able to flexibly adapt to moderate environment disturbances. This suggests two considerations when designing a gait reward function. First, the reward must be specific enough to produce the desired gait characteristic when optimized. Second, to account for the fact that there is uncertainty about the exact details of a gait in the context of specific terrain and dynamic conditions, the reward should not be overly constraining.

The common use of reference trajectories to specify gait-specific rewards, e.g., \cite{Zhang2018, Starke2020localmotion, xie2019iterative, Siekmann2020, Peng2018deepmimic}, partly addresses the first consideration above, but mostly ignores the second. In particular, a reference trajectory only captures a small part of the variation needed to realize a gait characteristic under varying conditions. Thus, attempting to adhere to such a trajectory can prevent learning a characteristic gait that is more robust and/or efficient, not to mention that deriving feasible reference trajectories for a particular desired gait can be very challenging in the first place. 

Reference-free approaches to specifying reward functions for locomotion are often highly underspecified, for example, those used in the OpenAI Gym \cite{openaigym} locomotion benchmarks. With this starting point, achieving a specific gait characteristic requires iterations of heuristic reward-function adjustments, based on observed RL performance, until arriving at a desired behavior. This approach can be tedious when it works and is unreliable as a general framework. Other reference-free approaches structure the reward around a specific type of locomotion behavior \cite{hwangbo2019learning} without being easily extended to other behaviors. 

\begin{figure}[t!]
\centering
\centerline{\includegraphics[width=0.54 \textwidth]{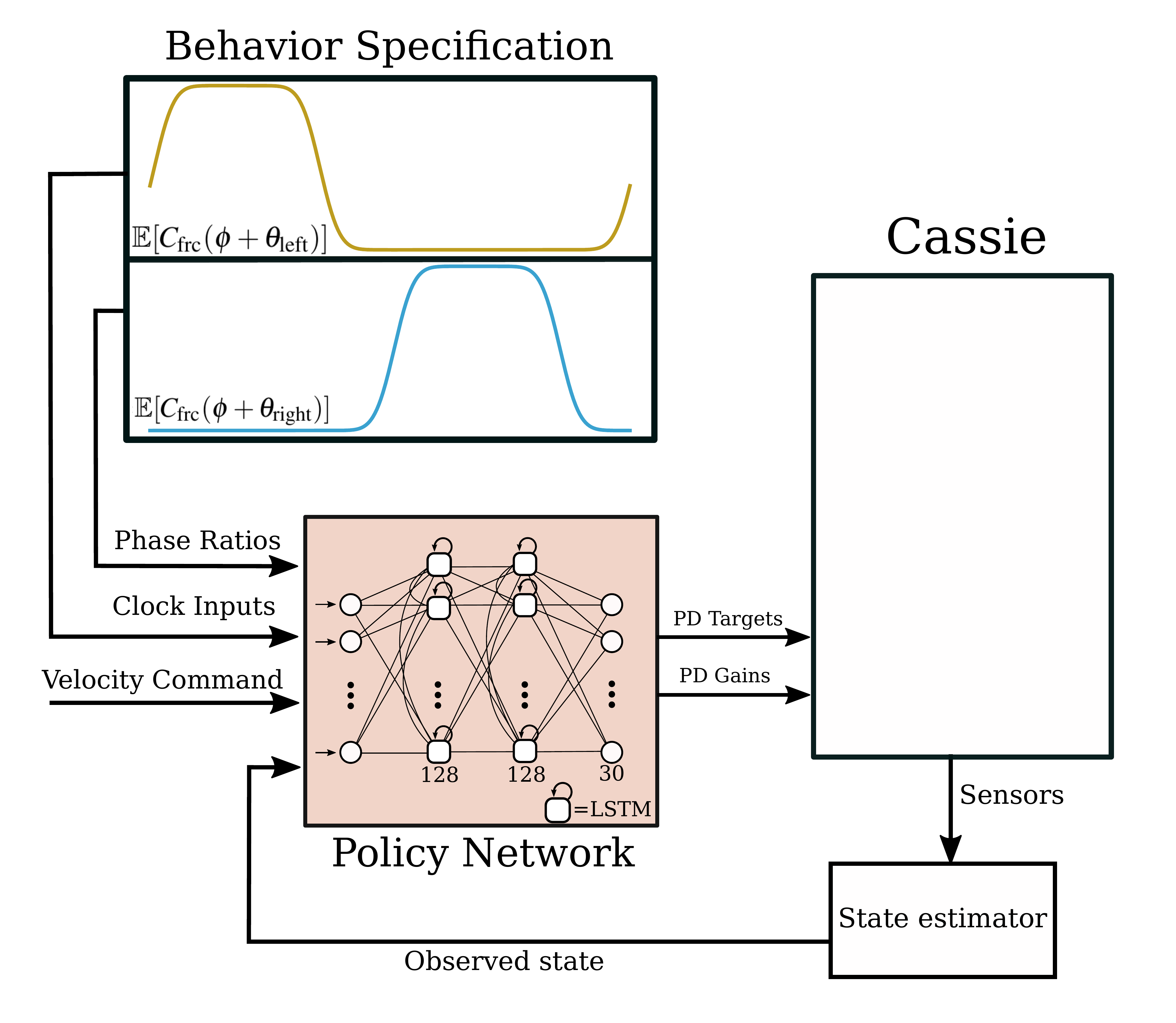}}
\caption{In this work, we present a reward design framework which makes it easy to learn policies which can stand, walk, run, gallop, hop, and skip on hardware. We condition the reward function based on a number of gait parameters, and also provide these parameters to the LSTM policy, which outputs PD joint position targets and PD gains to the robot.}
\label{fig:title_fig}
\end{figure}
\begin{figure*}[!htbp]
\centering
\includegraphics[width=1.0 \textwidth]{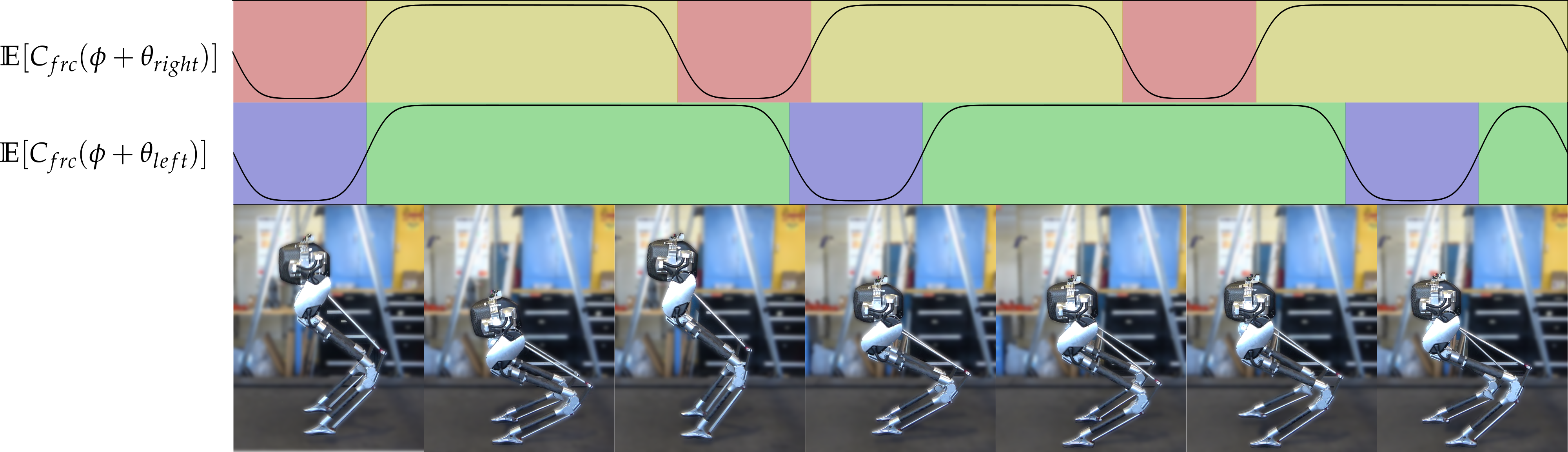}
\caption{A series of images showing a neural network policy controlling Cassie and continuously transitioning from hopping to galloping to walking. We present a simple reward design paradigm which makes use of probabilistic intervals to apply cost functions at specific times, allowing policies to learn all common bipedal gaits exhibited by animals in nature.}
\label{fig:cassie_continuous}
\end{figure*}

The first contribution of our work is to present a principled framework for designing reward functions that can naturally capture all of the periodic bipedal locomotion gaits. We are motivated by the fact that all common bipedal gaits can be defined by periodic \textit{swing phases} (foot swinging in the air) and  \textit{stance phases} (foot planted on the ground) for each foot \cite{gan2018all}.
A fundamental distinction between swing and stance phases is the complementary presence and absence of foot forces and foot velocities for a given foot.
We can create principled reward functions based on this observation by using the magnitudes of foot forces and velocities such that during a swing phase, forces are penalized while velocities are allowed, prompting the policy to learn to lift the foot.
Thus, our framework describes gaits as a sequence of periodic phases, each of which rewards or penalizes a particular measurement of the physical system. Our hypothesis is that this framework will allow for a more natural specification of reward functions that sufficiently constrain RL to learn the desired gait characteristics, while allowing for flexible adaptation to specific environmental disturbances.

Our second contribution is to demonstrate this framework for sim-to-real RL of all common bipedal gaits, including walking, running, galloping, skipping, and hopping, without using a motion capture dataset or reference trajectories.
We train policies for each of these behaviors in simulation and successfully demonstrate them on hardware. Further, by providing the framework's gait parameters as a command input to the policy, we are able to learn a multi-gait policy which can hop, gallop, run, and walk.

\section{Background}

The practice of synthesizing locomotion behaviors has been studied in a variety of fields.
In character animation, kinematics-based approaches (i.e., those using motion capture data) are commonly used to synthesize locomotion \cite{levine2012continuous}  \cite{safonova2004synthesizing} \cite{wieber2006online}, with some newer approaches using deep learning  \cite{Zhang2018} to train neural networks to solve problems like adapting realistically to varying ground geometries \cite{Holden2017} \cite{starke2019neural} or blending several types of actions into one realistic body motion \cite{Starke2020localmotion}.
Physics-based character animation, wherein the pose of a character is controlled through the application of torques and motions are simulated using a physics engine, has also come into prominence in recent years \cite{2013-SCA-diverse}, though is notably more difficult to use effectively for complicated motions \cite{geijtenbeek2011interactive} when compared to kinematics-based methods.

Approaches which use reinforcement learning to synthesize locomotion bear heavy similarities to the aforementioned methods used in character animation.
Much recent work uses trajectory-matching reward functions to train policies to imitate some reference trajectory (akin to a motion capture) while subjecting the policies to simulated physics, resulting in policies that behave realistically while imitating some reference trajectory \cite{bergamin2019drecon} \cite{Peng2018deepmimic} \cite{xie2019iterative} \cite{peng2016terrain} \cite{Siekmann2020}.
Reinforcement learning has also been used for synthesizing locomotion without the use of reference trajectories, but these approaches often place little emphasis on subjective quality of behaviors, resulting in behaviors which maximize some objective while producing policies which are often inefficient, infeasible or unsafe to execute in the real world, and not usually visually pleasing \cite{openaigym} \cite{deepmindcontrol}.
Methods which do prioritize physically realistic behavior without using a reference trajectory exist \cite{haarnoja2018learning} \cite{Tan-RSS-18} \cite{hwangbo2019learning}, but their reward functions are specific to single behaviors and are not trivial to extend to other behaviors.

%In this work we present a principled reward framework that can easily and consistently produce a multitude of distinct locomotion behaviors using deep reinforcement learning, without the use of the reference trajectories that previous approaches have required to synthesize realistic locomotion behavior. In addition, the framework we present is more extensible than the complex heuristic-based reward functions others have presented in the past. While we show results on hardware specifically for the application of this framework to bipedal locomotion, it can be extended to any number of appendages, to behaviors outside of legged locomotion, and even to aperiodic behaviors.

%\section{Learning Bipedal Gaits with a Probabilistic Behavior Framework}
\section{Learning Bipedal Gaits with Periodic Reward Composition}
\label{sec:framework}

\subsection{Reinforcement Learning Framework}
We formulate our problem in the framework of reinforcement learning (RL) \cite{Sutton2018}, for which we assume basic familiarity. The world is modeled as a discrete-time Markov Decision Process (MDP) with continuous state space $S$, continuous action space $A$, transition function $T(s,a,s')$, and reward function $R(s,t)$. Here $T(s,a,s')$ gives the probability density over the next state $s'$ after taking action $a$ in state $s$, and $R(s,t)$ gives the non-stationary reward for being in state $s$ at time step $t$.

A control policy is a possibly stochastic mapping $\pi(a\;|\;s)$ from states to actions, which dictates behavior. Given a policy the expected $T$-horizon discounted return is given by $J(\pi)=\mathbb{E}\left[\sum_{t=0}^{T}\gamma^t R(S_t,t)\right]$, where $\gamma \in [0,1]$ is a discount factor and $S_t$ is a random variable representing the state at time $t$ when following policy $\pi$ under transition dynamics $T$. The goal of RL is to learn a policy $\pi$ that maximizes $J(\pi)$ based on trial-and-error training experience in the world. In this work, we follow a sim-to-real RL paradigm where training is done in simulation to identify a policy, which is then used in the real-world.

\subsection{Periodic Reward Composition}
Since our framework is targeted toward periodic behaviors, we index time via a \textit{cycle time} $\phi$ variable, which repeatedly cycles over a normalized time period of $[0,1]$ at discrete time steps rather than increasing monotonically. Accordingly, the non-stationary reward function $R(s,\phi)$ is periodic and defined in terms of $\phi$ rather than absolute time.

As motivated in the introduction, our framework specifies rewards in terms of compositions of rewards on periodic intervals. We define the reward as a biased sum of $n$ \textit{reward components} $R_i(s,\phi)$, where each component $R_i(s,\phi)$ captures a desired characteristic of the gait during a particular phase.
$$R(s,\phi)=\beta + \sum_i R_i(s,\phi)$$
Each reward component $R_i(s,\phi)$ is a product of a \emph{phase coefficient} $c_i$, a \emph{phase indicator} $I_i(\phi)$, and a real-valued \emph{phase reward measurement} $q_i(s)$ (e.g. norm of a foot force).
%
%\jas{Where each component is a product of a \emph{phase coefficient} $c_i$, a binary random variable \emph{phase indicator} $I_i(\phi)$, and a positive scalar \emph{phase reward measurement} $q_i(s)$ (e.g. norm of a foot force).}
%
\begin{align}
    R_i(s,\phi) &= c_i \cdot I_i(\phi) \cdot q_i(s) \nonumber% \\
%    & \text{where } q_i(s) \in \mathbb{R}, q_i(s) \geq 0 \nonumber
\end{align}
The phase coefficient $c_{i}$ is a scalar whose sign determines the effect that the phase measurement $q_i(s)$ has on the total reward during cycle times when the reward component is active. 
The phase indicator function $I_i(\phi)$ is a binary-valued random variable denoting whether the target phase is active or not at cycle time $\phi$.
In this work, the distribution of each $I_i$ is described by random variables $A_i$ and $B_i$ representing the start and end times of the period respectively. Since $A_i$ and $B_i$ represent intervals on a cycle, we use Von Mises distributions (approximations of the wrapped Normal distribution) described by the parameter tuple $(a_i, b_i, \kappa)$, which gives the means $a_i$ and $b_i$ and shared variance parameter $\kappa$.
The distribution of the binary phase indicator $I_i(\phi)$ is then simply:
%
%
%The phase coefficient $c_{i}$ is a scalar whose sign determines the effect on the reward signal for the phase reward measurement $q_i(s)$:
%
% %  BELOW IS FOR OLD "Reward Components"
% \begin{table}[h]
% %\color{Blue}
% \centering
% \begin{tabular}{|l|l|}
% \hline
% $c_{ij}$ & Effect on reward for $q_i(s)$ during phase $j$\\ \hline
% $< 0$       & penalize (negative reward)    \\ \hline
% $= 0$       & ignore (no reward)            \\ \hline
% $> 0$       & incentivize (positive reward) \\ \hline
% \end{tabular}
% \caption{Influence of the sign of phase coefficients on the total reward. % TODO: add more to this caption, not sure what though
% }
% \label{table:coefficients}
% \end{table}
%
% The phase indicator function $I_i(\phi)$ is a binary-valued random variable denoting whether the target phase is active at time $\phi$.
% For this work, the distribution of each $I_i$ is parameterized by a tuple $(a_i, b_i, \kappa)$ where the start time $a_i$, end time $b_i$, and scale parameter $\kappa$ specify two random variables $A_i$ and $B_i$ with Von Mises (wrapped normal) distributions.
% The distribution of the binary random phase indicator function $I_i(\phi)$ is then simply:
%
\begin{align}
\label{eqn:von_mises}
  P\left(I_i(\phi)=x\right) =&
  \begin{cases}
       P(A_{i} < \phi < B_{i}) & \text{if $x=1$} \\
       1 - P(A_{i} < \phi < B_{i}) & \text{if $x=0$}
  \end{cases} \\ \nonumber \\
  \text{where } &P(A_{i} < \phi < B_{i}) = P(A_i < \phi)(1 - P(B_i < \phi)) \nonumber \\
& A_i \sim \boldsymbol{\Phi}(2 \pi a_{i}, \kappa) \text{ and } B_i \sim \boldsymbol{\Phi}(2 \pi b_{i}, \kappa) \nonumber\\
& \boldsymbol{\Phi} \text{ is the Von Mises distribution } \nonumber
\end{align}

Note that this formulation allows for uncertainty about the exact start and end of each phase to be captured via the variance parameter $\kappa$ of the Von Mises distributions. This has a smoothing effect on the reward function at phase boundaries, which we have found to usefully encourage more stable and consistent learning. While we could directly use this probabilistic reward function $R(s,\phi)$ for RL training by sampling from the reward distribution at each step, we instead apply RL to the deterministic expectation of $R(s,\phi)$. 

\begin{align}
% \mathbb{E}\left[\mathbf{R}(s,\phi)\right] &= \sum_i^n c_i \cdot \mathbb{E}[I_i(\phi)] \cdot q_i(s) + \beta
\mathbb{E}\left[R(s,\phi)\right] &= \sum_i^n c_i \cdot \mathbb{E}[I_i(\phi)] \cdot q_i(s) + \beta \nonumber
\end{align}

%This decreases the variance of the learning algorithm while not impacting the theoretical optimum policy. In particular,
Due to the linearity of expectations, for any policy $\pi$ the expected cumulative return $J(\pi)$ is the same regardless of whether we use stochastic rewards or their expectations.

\begin{figure}[t]
\centering
\centerline{\includegraphics[width=0.4 \textwidth]{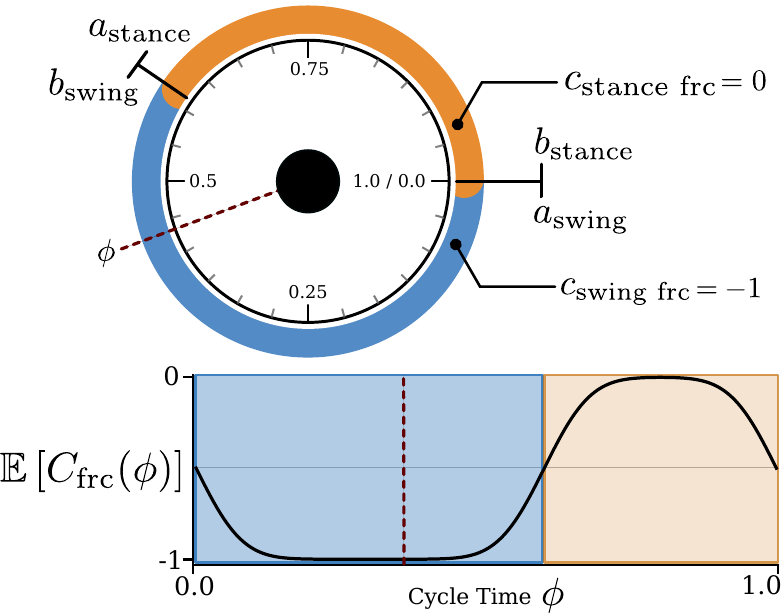}}
\caption{The circular intervals of two phases, swing and stance, are shown on a polar plot. The expected sum of the phase indicators and phase coefficients for foot force $C_{\text{frc}}(\phi)$ is shown below.}
\label{fig:framework_explanation}
\end{figure}

\subsection{Describing Bipedal Gaits}

For simplicity, we begin by describing repeatedly lifting and placing a single foot, or equivalently, cycling between swing and stance phases with our framework.
As our phase reward measurements, we select the norm of foot force $q_\text{frc}(s)$ and the norm of foot velocity $q_\text{spd}(s)$.
During the swing phase we want to penalize foot forces and ignore foot velocities, so we choose
$c_\text{swing frc}=-1$ and $c_\text{swing spd}=0$. %, per Table \ref{table:coefficients}
Similarly, we choose $c_\text{stance spd}=-1$ and $c_\text{stance frc}=0$ to penalize foot velocities and ignore foot forces during the stance phase.
We constrain the swing and stance phases to follow immediately after one another and together last the entire cycle time by defining a ratio $r \in (0,1)$ and setting the intervals for both phases such that the swing phase lasts length $r$, while the stance phase lasts length $1-r$ and starts directly afterwards.
Finally, by choosing a common scale $\kappa$, we can define the indicator functions $I_\text{frc}(\phi)$ and $I_\text{spd}(\phi)$ by using Equation \ref{eqn:von_mises}.
%We also want these phases to follow immediately after one another, so we set $a_\text{frc}=0$, $b_\text{frc}=a_\text{spd}$, and $b_\text{spd}=1$.
%We use these measurements in 
%by defining random variables $C_\text{frc}(\phi)$ and $C_\text{spd}(\phi)$, where each component has two phases, with coefficients:

%\begin{align*}
%c_{\text{frc}_0} &= -1 & c_{\text{spd}_0} &= 0 \\
%c_{\text{frc}_1} &= 0 & c_{\text{spd}_1} &= -1
%\end{align*}

For convenience, we define $C_{\text{frc}}(\phi)$ and $C_{\text{spd}}(\phi)$ as the expected sum of the product of all the phase indicators and phase coefficients for the swing and stance phases:
\begin{equation*}
\begin{aligned} 
\displaystyle
\mathbb{E} \left[C_{\text{frc}}(\phi)\right] & = & & c_\text{swing frc} \cdot \mathbb{E}[I_{\text{swing frc}}(\phi)] \\
& & + & c_\text{stance frc} \cdot \mathbb{E}[I_{\text{stance frc}}(\phi)] \\
\mathbb{E} \left[C_{\text{spd}}(\phi)\right] & = & & c_\text{swing spd} \cdot \mathbb{E}[I_{\text{swing spd}}(\phi)] \\
& & + & c_\text{stance spd} \cdot \mathbb{E}[I_{\text{stance spd}}(\phi)] \\
% \mathbb{E} \left[\mathbf{R}_{\text{unipedal}}(s,\phi)\right] = && c_\text{swing frc} \cdot \mathbb{E}[I_{\text{swing frc}}(\phi)] \cdot q_{\text{frc}}(s)\\
% &+&  c_\text{stance frc} \cdot \mathbb{E}[I_{\text{stance frc}}(\phi)] \cdot q_{\text{frc}}(s)\\
% &+& c_\text{swing frc} \cdot \mathbb{E}[I_{\text{swing spd}}(\phi)] \cdot q_{\text{spd}}(s)\\
% &+& c_\text{stance spd} \cdot \mathbb{E}[I_{\text{stance spd}}(\phi)] \cdot q_{\text{spd}}(s)\\
\end{aligned}
\end{equation*}

Refer to Fig.\ref{fig:framework_explanation} for a visual explanation of the phase start and end times, $\phi$, and the expected value of $C_\text{frc}$. For more complicated behaviors, we find that visualizing $C_{\text{frc}}(\phi)$ can be useful for understanding how the reward function changes over the cycle to guide the learning of a particular behavior.

Putting the force and speed components together, the expected overall reward for repeatedly lifting and placing a single foot is
\begin{equation*}
\begin{aligned}
% \mathbb{E} \left[\mathbf{R}_{\text{unipedal}}(s,\phi)\right] & = & & \mathbb{E} \left[C_{\text{frc}}(\phi)\right] \cdot q_{\text{frc}}(s) \\
% & & + & \mathbb{E} \left[C_{\text{spd}}(\phi)\right] \cdot q_{\text{spd}}(s)
\mathbb{E} \left[R_{\text{unipedal}}(s,\phi)\right] & = & & \mathbb{E} \left[C_{\text{frc}}(\phi)\right] \cdot q_{\text{frc}}(s) \\
& & + & \mathbb{E} \left[C_{\text{spd}}(\phi)\right] \cdot q_{\text{spd}}(s)
\end{aligned}
\end{equation*}
%\subsection{Bipedal Example}

% 4. Introduce \emph{cycle offset}, multiple feet
Bipedal gaits are behaviors where the left and right feet both follow the same sequence of phases described above, but offset relative to each other in phase time.
For instance, in a walking behavior the timings of the swing and stance phases are shifted apart by half of the period length (one leg in swing, the other in stance), while a hopping behavior is one where both feet synchronously enter the swing and stance phases.
To expand the simple behavior of lifting and placing a single foot into the full spectrum of bipedal gaits, we need only introduce two \emph{cycle offset} parameters $\theta_\text{left}, \theta_\text{right}$ which define the exact timing shift between the identical sequence of behavioral phases for the left and right feet, and differentiate between the norms of left and right foot forces, $q_\text{left frc}(s),\ q_\text{right frc}(s)$ and norms of left and right foot velocities $q_\text{left spd}(s),\ q_\text{right spd}(s)$.
The expected overall reward for a bipedal behavior is
\begin{equation}
\begin{aligned} 
\displaystyle
%\mathbb{E}[\mathbf{C}_{\text{bipedal}}(s,\phi)] & = & &\alpha_0 \cdot \mathbb{E}[C_{\text{frc}}(\phi+\theta_\text{left})] \cdot K(q_{\text{left frc}}(s)) \\
%& & + &\alpha_1 \cdot \mathbb{E}[C_{\text{frc}}(\phi+\theta_\text{right})] \cdot K(q_{\text{left frc}}(s))\\
%& & + &\alpha_2 \cdot \mathbb{E}[C_{\text{spd}}(\phi+\theta_\text{left})] \cdot K(q_{\text{left spd}}(s)) \\
%& & + &\alpha_3 \cdot \mathbb{E}[C_{\text{spd}}(\phi+\theta_\text{right})] \cdot K(q_{\text{left spd}}(s))\\
% \mathbb{E}[\mathbf{R}_{\text{bipedal}}(s,\phi)] & = & & \mathbb{E}[C_{\text{frc}}(\phi+\theta_\text{left})] \cdot q_{\text{left frc}}(s) \\
% & & + & \mathbb{E}[C_{\text{frc}}(\phi+\theta_\text{right})] \cdot q_{\text{right frc}}(s)\\
% & & + & \mathbb{E}[C_{\text{spd}}(\phi+\theta_\text{left})] \cdot q_{\text{left spd}}(s) \\
% & & + & \mathbb{E}[C_{\text{spd}}(\phi+\theta_\text{right})] \cdot q_{\text{right spd}}(s)\\
\mathbb{E}[R_{\text{bipedal}}(s,\phi)] & = & & \mathbb{E}[C_{\text{frc}}(\phi+\theta_\text{left})] \cdot q_{\text{left frc}}(s) \\
& & + & \mathbb{E}[C_{\text{frc}}(\phi+\theta_\text{right})] \cdot q_{\text{right frc}}(s)\\
& & + & \mathbb{E}[C_{\text{spd}}(\phi+\theta_\text{left})] \cdot q_{\text{left spd}}(s) \\
& & + & \mathbb{E}[C_{\text{spd}}(\phi+\theta_\text{right})] \cdot q_{\text{right spd}}(s)\\
\end{aligned}
\label{eqn:bipedal_eq}
\end{equation}

\begin{figure}[ht]
\centering
\centerline{\includegraphics[width=0.45 \textwidth]{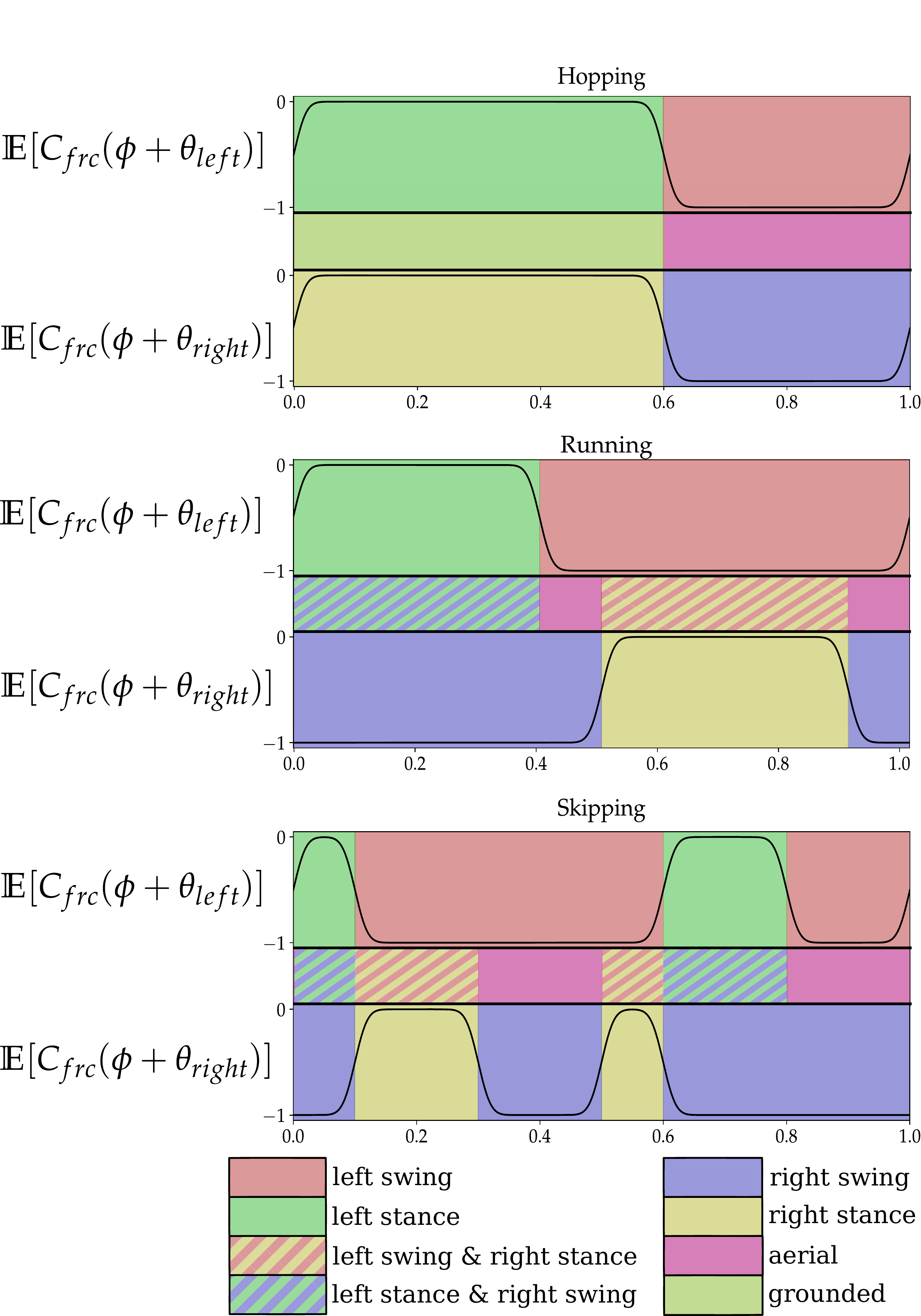}} % max width=0.45
\caption{Plot of the expected value of the force phase coefficient for each foot for some example bipedal gaits. In hopping there is a $|\theta_\text{left} - \theta_\text{right}| \approx 0$ shift between the left and right feet. For walking and running, $|\theta_\text{left} - \theta_\text{right}| \approx 0.5$, with transitional shifts resulting in galloping. Four phases are necessary for describing skipping, as well as a $|\theta_\text{left} - \theta_\text{right}| \approx 0.5$ shift between the left and right feet. The shaded region between the phase coefficient plots show the hybrid phases from combining the left and right foot behaviors together.}
\label{fig:example_gaits}
\end{figure}

Introducing $\theta_\text{left}, \theta_\text{right}$ for two phase behaviors allows us to define reward functions for walking, galloping, and hopping. Additionally, by using four phases rather than two in each component, we can derive a skipping reward function, as shown in Fig. \ref{fig:example_gaits}.

\section{Method}

\textbf{Network Architecture and Action Space:} For all policies, we use a Long Short-Term Memory neural network \cite{hochreiter1997long} with two recurrent hidden layers of size 128 each, and a simple linear output projection of size 30, corresponding to 10 desired joint positions, and 10 sets of PD gains, as seen in Fig. \ref{fig:title_fig}. The desired joint positions are added to a set of constant offsets corresponding to a neutral standing position, such that an output vector of zeroes results in a standing pose. Similarly, the P gains and D gains are also summed with a fixed 'neutral' set of gains. The policy is evaluated at 40Hz, and the robot's PD controllers are run at 2000Hz. A similar system is used in \cite{xie2019iterative} and \cite{Siekmann2020}, though without PD gain deltas.

\textbf{State Space:} To prevent the reinforcement learning environment from becoming a non-stationary Markov decision process, some information about the periodic reward function must be present in the state provided to the policy. Specifically, the policy must receive some sort of encoding of the current cycle time $\phi$, an encoding of the cycle offset parameters, $\theta_\text{left},\theta_\text{right}$, and information about the start and end times of the phases of the phase reward components.

In order to encode the current cycle time $\phi$ and the cycle offset parameters $\theta_\text{left},\theta_\text{right}$, we condition the policies on two clock inputs: 
\[
p = \left\{\sin\left(\frac{2 \pi (\phi + \theta_\text{left})}{L}\right), \sin\left(\frac{2 \pi (\phi + \theta_\text{right})}{L}\right)\right\}
\]
where $L$ is the number of discrete timesteps in the entire cycle. To encode information about the start and end times of the phases into the state, we derive a vector of \emph{ratios} from the sequence of phase timings; each ratio represents the proportion out of the total period that a phase occupies. For instance, a phase $j$ with start time $a_{i, j=0.3}$ and end time $b_{i,j}=0.7$ should occupy 40\% of the total period time (plus or minus some uncertainty); thus, we can derive a ratio $r_j=0.4$. Each phase's ratio is calculated and provided to the policy.

% TERMINOLOGY: Clock input duty cycle offsets = cycle offsets?
%Since the reward function distinguishes between Cassie's two legs with two separate cycle offsets, or duty cycle offsets, $\theta_\text{left},\theta_\text{right}$, it is not sufficient to provide a clock input tracking a global phase as in previous efforts in applying deep reinforcement learning to Cassie \cite{Siekmann2020} \cite{xie2019iterative} or in character animation approaches \cite{Holden2017} \cite{Zhang2018}. Instead, the policy must be provided with local clock inputs tracking the position in the duty cycle for each leg. % better phrasing?

Thus, the policy's input consists of:
\[ X_t =  \left\{
\begin{array}{ll}
      \hat{q},\ \hat{\dot{q}} & \text{robot state} \\
      \smallskip
      \dot{x}_{\text{desired}},\dot{y}_{\text{desired}} & \text{desired velocity} \\
      r, p & \text{phase ratios and clock inputs} \\
\end{array} 
\right. \]

Where $\hat{q},\hat{\dot{q}}$ are estimates of the pelvis orientation, rotational velocity, joint positions and joint velocities. $\dot{x}_{\text{desired}}$ and $\dot{y}_{\text{desired}}$ are speed commands randomized during training, and manually controlled by the user during evaluation.

%Kinematics-based character animation controllers parameterized by neural networks generally use a global phase variable to track the current time in the motion cycle and sometimes to perform a blending operation or interpolation between multiple sets of neural networks, each performing some subtask \cite{Holden2017} \cite{Zhang2018}. While we do use a global phase variable In our work however, we 

\textbf{Reward Formulation:}
% The cost components we use to train the policy to learn specific bipedal gaits across all policies are of the form,
% \begin{equation*}
% \begin{aligned} 
% \displaystyle
% \mathbb{E}[\mathbf{C}_{\text{behavior}}(s,\phi)] & = &\alpha_0 & \cdot \mathbb{E}[C_\text{frc}(\phi + \theta_\text{left}] \cdot K(q_\text{left frc}(s)) \\ 
% & + &\alpha_1 &\cdot \mathbb{E}[C_\text{frc}(\phi + \theta_\text{right}] \cdot K(q_\text{right frc}(s))  \\
% & + &\alpha_2 &\cdot \mathbb{E}[C_\text{spd}(\phi + \theta_\text{left}] \cdot K(q_\text{left spd}(s)) \\ 
% & + &\alpha_3 &\cdot \mathbb{E}[C_\text{spd}(\phi + \theta_\text{right}] \cdot K(q_\text{right spd}(s))  \\ \smallskip
% \end{aligned}
% \end{equation*}
% We use the set of reward components from $\mathbf{R}_{\text{bipedal}}$ defined in equation \ref{eqn:bipedal_eq} to learn specific bipedal gaits across all of our policies.
We use the set of reward components from $R_{\text{bipedal}}$ defined in Equation \ref{eqn:bipedal_eq} to learn single-gait policies. In addition, we also use a variety of auxiliary cost components to further constrain the path of viable exploration towards stable locomotion and facilitate successful sim-to-real transfer as well as command the policy to match a desired orientation, forward speed, and sidespeed. These rewards do not need to vary as a function of time, and can be seen as a special case of the reward component framework, wherein the random variable $c_i$ has only one phase, and thus one possible value (in this case, $-1$, to show that we wish to penalize these quantities $q_i$ for the entire period).

\begin{equation*}
\begin{aligned} 
\displaystyle
% \mathbf{R}_{\text{cmd}}(s) & = & & (-1) \cdot q_{\dot{x}}(s)  \\
% & + & &(-1) \cdot q_{\dot{y}}(s)  \\
% & + & & (-1) \cdot q_\text{orientation}(s)  \\
% \mathbf{R}_{\text{smooth}}(s) & = & & (-1) \cdot q_\text{action diff}(s)  \\
% & + & & (-1) \cdot q_\text{torque}(s)  \\
% & + &  & (-1) \cdot q_\text{pelvis acc}(s)  \\
R_{\text{cmd}}(s) & = & & (-1) \cdot q_{\dot{x}}(s)  \\
& + & &(-1) \cdot q_{\dot{y}}(s)  \\
& + & & (-1) \cdot q_\text{orientation}(s)  \\
R_{\text{smooth}}(s) & = & & (-1) \cdot q_\text{action diff}(s)  \\
& + & & (-1) \cdot q_\text{torque}(s)  \\
& + &  & (-1) \cdot q_\text{pelvis acc}(s)  \\
\end{aligned}
\end{equation*}

\begin{equation}
% \mathbb{E}[\mathbf{R}(s,\phi)] =  \mathbb{E}[\mathbf{R}_{\text{bipedal}}(s,\phi)] + \mathbf{R}_{\text{smooth}}(s)] + \mathbf{R}_{\text{cmd}}(s) + \beta
\mathbb{E}[R(s,\phi)] =  \mathbb{E}[R_{\text{bipedal}}(s,\phi)] + R_{\text{smooth}}(s)] + R_{\text{cmd}}(s) + \beta
\end{equation}

Where $q_{\dot{x}}$ and $q_{\dot{y}}$ are measures of error between the commanded velocity and the actual pelvis velocity, and $q_\text{orientation}$ is the negative exponent of quaternion difference between the pelvis orientation and an orientation which faces straight, which can be used to change the heading of the robot. $q_\text{pelvis acc}$ is a cost for aggressive pelvis motion, which helps reduce noise in the state estimator, while $q_\text{action diff}$ is a cost for aggressive actions and $q_\text{torque}$ is an overall joint torque cost to encourage efficient motions. For general policies which can learn and transition between a continuum of bipedal gaits, we specify additional reward terms which are discussed in Section \ref{sec:results}.

\textbf{Dynamics Randomization:}
In order to facilitate successful sim-to-real transfer, we use dynamics randomization \cite{peng2017sim2real} \cite{tan2018simtoreal} to expose the policies to a wide variety of possible real-world dynamics. Specifically, we randomize the execution rate of the policy, the mass and damping of the joints, the friction of the ground, the slope of the ground, and joint position encoder noise. Details on the ranges and distributions used to randomize these parameters can be found in Table \ref{table:dynamicsrand}. These parameters are randomized at the beginning of every rollout during training and remain constant throughout each rollout.

\begin{table}[!h]
\centering
\begin{tabular}{|l|l|l|}
\hline
Parameter & Unit & Range \\ \hline
 Joint damping                      & Nms/rad   &   $[0.3, 4.0] \times \text{default values}$ \\ \hline
 Joint mass                         & kg        &   $[0.5, 1.5] \times \text{default values}$     \\ \hline
 Ground Friction                    & --   &   $[0.35, 1.1] $     \\ \hline
 Ground Slope                       & rad  &   $[-0.03, 0.03]$     \\ \hline
 Joint Encoder Offset               & rad  &   $[-0.05, 0.05]$     \\ \hline
 %Execution Rate               & Hz  &   $[36, 44]$     \\ 

 \hline

\end{tabular}
\caption{The ranges for randomization of several dynamics parameters during training. We use a uniform distribution over the given ranges for all listed parameters.}
\label{table:dynamicsrand}
\end{table}

\textbf{Proximal Policy Optimization:} To train our policies, we use a common model-free reinforcement learning algorithm known as Proximal Policy Optimization (PPO) \cite{schulman2017proximal}. More specifically, we use the recurrent adaptation described in \cite{Siekmann2020}, which samples batches of trajectories from a replay buffer, rather than batches of individual timesteps. In our case, we use a recurrent critic with exactly the same dimensions as the actor with the exception of the final linear output vector, which in the case of the critic is a scalar. We use a fixed exploration action noise, with standard deviation $e^{-1}$. In addition, we incorporate a mirror loss term \cite{2019-MIG-symmetry} into our policy gradient algorithm with the aim of achieving more symmetric locomotion behavior. 
%The mirror loss operates on the norm of the difference in action taken when the policy is given the same state with joint positions and clock inputs from each leg flipped to the other side, and the pelvis orientation is mirrored; effectively, the entire state input is reflected across the global sagittal plane. Minimizing this loss has the effect of reducing the asymmetry in actions taken by the policy when it is given two states which are identical except they are mirrored across the sagittal plane.

\section{Experimental Results}
\label{sec:results}
%\begin{figure}[b!]
%\centering
%\includegraphics[width=0.48 \textwidth]{Figures/skipping_standalone.pdf}
%\caption{A series of images showing a single-gait learned policy controlling Cassie for the task of skipping, with a graph of the expectation of the random variable $C_\text{force}$ used in the probabilistic reward underneath. When the expectation of the random variable is close to -1, the policy is punished for applying foot forces, causing it to learn to lift its foot.}
%\label{fig:cassie_discrete}
%\end{figure}

{\bf Training Details.} Policies were trained using PPO with a batch size of 32 trajectories of up to 300 timesteps each, a learning rate of $0.0001$ for both the actor and critic, a replay buffer of 50,000 samples, and 4 epochs per iteration. Training was terminated after 150,000,000 samples, which took between 24 and 36 hours per policy. We use the \textit{cassie-mujoco-sim} \cite{AgilityRobotics2018} simulator, based on MuJoCo \cite{todorov2012mujoco}, to simulate Cassie during training.

{\bf Single-Gait Policy Results.} We found that it was straightforward to use the probabilistic framework to train policies to learn standalone gaits, such as hopping, walking, running, or even skipping.
These behaviors can be learned simply by holding the values of ratios $r$ and cycle offset parameters $\theta_\text{left}, \theta_\text{right}$ to be constant throughout training 
(we refer the reader to Fig. \ref{fig:example_gaits} for examples of gait specifications).
Videos of single-gait policies which learn skipping hopping, and walking can be found in our submission video.

\textbf{Multi-Gait Policy Results} 
By keeping the cycle offsets fixed and varying the phase ratios over some range during training, we are able to train policies that can learn to hop with more or less time in the air, and policies that can transition from walking to running.
However, learning to transition between gaits by varying both the cycle offsets and phase ratios during training appears to be a challenging learning problem: policies which are trained in this fashion can end up asymmetrically walking instead of hopping, or learn other undesirable behaviors that resemble a fusion of all the different commanded gaits.
To avoid these issues, we introduce additional reward terms called \textit{transition penalties} to further distinguish each of the desired behaviors from each other during training.
Transition penalties are behavior-specific costs which are only active when specific behaviors are being commanded.
For hopping, which is commanded when cycle offsets are very close to each other, we activate a cost for constraining the feet positions to be close together.
Similarly, by adding another transition penalty for standing, we can train a generic controller to transition between all steady state two-beat gaits and standing in place.
Full details of the reward function for generic controllers are described in detail in the Appendix.

\textbf{Outdoor Experiments}
We demonstrate the robustness of policies learned with our approach in a variety of outdoor experiments, shown in our submission video \footnote{\href{https://youtu.be/4DnxV9lko\_U}{youtu.be/4DnxV9lko\_U}}.
Given that the robot is effectively blind to the external world, the behaviors we observe in response to disturbances are surprisingly robust. % how do i somehow hint at we're approaching limits of blind walking robustness
We show the multi-gait policy hopping on and off a sidewalk, walking with one foot elevated on a curb, and running over small bumps in its path.
The policy is also shown smoothly transitioning between all of the 2-beat gaits while moving forward on a crowned road, and while turning in a small circle on a turf field.
We also show a multi-gait policy descending down 5 steps of stairs while in a running gait.
In the staircase and sidewalk tests, we observe the foot slipping off of small ledges and being compliant to the slope of the ground.
This indicates a priority on force profile, as opposed to position only.

\begin{figure}[t]
\centering
\centerline{\includegraphics[width=0.5 \textwidth]{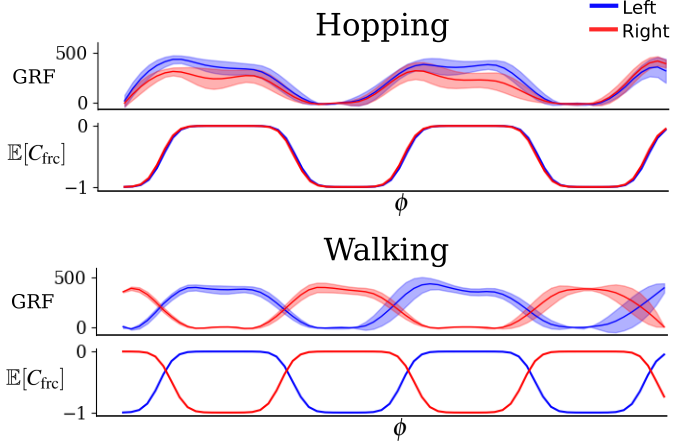}}
\caption{Comparison between the mean measured simulation ground reaction forces (GRFs) in newtons and the corresponding expected values of $C_\text{frc}$ over multiple policies. When the expectation is close to -1, the policies refrain from applying foot forces. When the expectation is close to 0, the policies apply foot forces.}
\label{fig:grf_comparison}
\end{figure}

\section{Conclusion}

In this work, we introduced a probabilistic reward framework which allows for learning of all of the common bipedal gait behaviors observed in animals. This framework requires no reference trajectories, enabling policies to explore a rich space of possible interaction with the world during training without artificial constraints to some arbitrary trajectory through space. We showed that not only are we able to train policies to learn all common bipedal gaits individually, including walking, running, hopping, galloping, and skipping, but also that we can learn all 2-beat behaviors on single policies and transition between them continuously, even while in motion. Some questions which remain unanswered include how easily this framework might be applied to the space of possible quadrupedal gaits, or other bipedal morphologies. A modified version of this framework could also be applied to aperiodic motions to learn one-off behaviors.

\section*{Appendix}\label{Appendix}

\subsection*{Full Reward Functions for Generic 2-Beat Policies}

When learning the continuum of 2-beat gaits, we find that policies often learn to stutter rather than hop with their feet together, which is not observed when training hopping standalone. To combat this, we introduce a \textit{hop symmetry} term, a cost which becomes active when the cycle offsets of both legs match closely and punishes large distances between the feet in the sagittal and transverse planes, $\text{err}_\text{sym}$,
\[
q_\textit{hop sym}(s) = 1-\exp(-\text{err}_\text{sym} \exp(-5 \left|sin(2\pi(\theta_\text{left} -\theta_\text{right}))\right|))
\]

To learn a standing behavior in addition to the rest of the 2-beat gaits, the reward function must be modified to reflect our wish for the policy to remain very still when commanded. First, we define the standing region of gait parameter space as one where the swing-to-stance phase ratio is close to 0. 

We define $\omega = (1 + \exp(-50(r_\text{swing} - 0.15)))^{-1}$, which is a coefficient close to one during normal locomotion (when the swing and stance ratios are in the range [0.35, 0.7], and close to zero during standing (where the swing phase ratio is close to zero).  Now we define an additional standing cost, which applies an additional action difference cost and a foot symmetry cost (similar to the hopping symmetry) when the command inputs are in the standing region.
\[
q_\text{standing cost}(s) = 1-\exp(-((\text{err}_\text{sym}) + 20q_\text{action diff}(s)))
\]
Now we define the full reward for a generic policy which includes standing:
\begin{equation*}
\begin{aligned} 
\displaystyle 
% \mathbb{E}[\mathbf{R}_{\text{multi}}(s,\phi)] &= & 0.400& \cdot \mathbb{E}[\mathbf{R}_{\text{bipedal}}(s,\phi)] \\
% &\ + &0.325&  \cdot\mathbb{E}[\mathbf{R}_{\text{cmd}}(s)] \\
% &\ + &0.100&   \cdot(\omega - 1)    \cdot q_\text{standing cost}(s)  \\ \smallskip
% &\ + &0.100&   \cdot(-1) \cdot q_\text{hop sym}(s)  \\ \smallskip
% &\ + &0.075& \cdot \mathbb{E}[\mathbf{R}_{\text{smooth}}(s)] \\
% &\ + &1& \\
\mathbb{E}[R_{\text{multi}}(s,\phi)] &= & 0.400& \cdot \mathbb{E}[R_{\text{bipedal}}(s,\phi)] \\
&\ + &0.300&  \cdot\mathbb{E}[R_{\text{cmd}}(s)] \\
&\ + &0.100& \cdot \mathbb{E}[R_{\text{smooth}}(s)] \\
&\ + &0.100& \cdot(\omega - 1)    \cdot q_\text{standing cost}(s)  \\ \smallskip
&\ + &0.100&  \cdot(-1) \cdot q_\text{hop sym}(s)  \\ \smallskip
&\ + &1& \\
\end{aligned}
\end{equation*}

\bgroup
\def\arraystretch{1.2}%  1 is the default, change whatever you need
\begin{table}[!h]
\centering
\begin{tabular}{|l|l|}
\hline
Quantity $q_i$ & Detail \\ \hline
$ q_\text{left/right frc}$ & $1-\exp(-\omega\|\text{raw\_foot\_frc}\|_2^2/100) $\\ \hline
$ q_\text{left/right spd}$ & $1-\exp(-2\cdot\omega\|\text{raw\_foot\_spd}\|_2^2) $ \\ \hline
$ q_{\dot{x}}$ & $1-\exp(-2\cdot\omega|\dot{x}_\text{desired} - \dot{x}_\text{actual}|) $ \\ \hline
$ q_{\dot{y}}$ & $1-\exp(-2\cdot\omega|\dot{y}_\text{desired} - \dot{y}_\text{actual}|) $ \\ \hline
$ q_\text{orientation}$ & $1-\exp(-3\cdot(1 - ((\text{quat}_\text{actual})^T(\text{quat}_\text{des}))^2) $ \\ \hline
$ q_\text{action diff}$ & $1-\exp(-5\cdot\|a_t - a_{t-1}\|) $  \\ \hline
$ q_\text{torque}$ & $1-\exp(-0.05\cdot\|\tau\|) $  \\ \hline
$ q_\text{pelvis acc}$ & $1-\exp(-0.10 \cdot (\|\text{pelvis}_\text{rot}\| + \|\text{pelvis}_\text{acc}\|) $ \\ \hline
\end{tabular}
\caption{}
\label{table:rewardspec}
\end{table}
\egroup

Where $\dot{x}_\text{actual}$ and $\dot{y}_\text{actual}$ are the current forward and lateral speeds of the pelvis. $\text{quat}_\text{actual}$ and $\text{quat}_\text{des}$ are the actual pelvis orientation and the desired pelvis orientation, in quaternion format. $a_t$ is the current timestep's action, and $a_{t-1}$ is the previous timestep's action. $\tau$ is the net torque applied to all joints across the robot. $\text{pelvis}_\text{rot}$ is the rotational velocity of the pelvis, and $\text{pelvis}_\text{acc}$ is the translational acceleration of the pelvis. Note that certain terms are multiplied by $\omega$, signifying that these terms should not be respected by the policy when $\omega\approx0$ (i.e., the policy is somewhere inside the standing region of the gait parameter space), to prevent the policy from (for example) attempting to match a desired speed while standing. The choice of $1 - \exp(-|x|)$ as a kernel function was influenced by a desire to have a reward bounded by 0 and 1.

\section*{Acknowledgements}

Thanks to Jeremy Dao, Helei Duan, and Kevin Green for stimulating conversations and help with hardware testing, and to Intel for providing access to the vLab academic compute cluster.


%%%%%%%%%%%%%%%%%%%%%%%%%%%%%%%%%%%%%%%%%%%%%%%%%%%%%%%%%%%%%%%%%%%%%%%%%%%%%%%%

%%%%%%%%%%%%%%%%%%%%%%%%%%%%%%%%%%%%%%%%%%%%%%%%%%%%%%%%%%%%%%%%%%%%%%%%%%%%%%%%

%%%%%%%%%%%%%%%%%%%%%%%%%%%%%%%%%%%%%%%%%%%%%%%%%%%%%%%%%%%%%%%%%%%%%%%%%%%%%%%%
% \pagebreak
%\bibliography

%\bibliographystyle{IEEEtran.bst}
%\bibliographystyle{IEEEtranN.bst}

%\bibliography{references.bib}

%\bibliographystyle{unsrtnat}
%\bibliography{references.bib}
%\bibliographystyle{plainnat}
%\bibliography{references.bib}
\printbibliography{}

\end{document}